\documentclass[letterpaper,10pt,conference]{ieeeconf}

\IEEEoverridecommandlockouts
\usepackage{cite}
\usepackage{amsmath,amssymb,amsfonts}
\usepackage{graphicx}
\usepackage{booktabs}
\usepackage{multirow}
\usepackage{url}
\usepackage{xcolor}
\usepackage{colortbl}
\usepackage{tabularx}
\newcolumntype{Y}{>{\centering\arraybackslash}X}

\definecolor{c1}{HTML}{D6ECF0}
\newcommand{\minisection}[1]{\par\noindent\textbf{#1}\enspace\ignorespaces}

\makeatletter
\let\NAT@parse\undefined
\makeatother
\usepackage[breaklinks,hidelinks,bookmarks=false]{hyperref}
\usepackage[capitalize,nameinlink]{cleveref}
\crefname{section}{Sec.}{Secs.}
\Crefname{section}{Sec.}{Secs.}
\crefname{subsection}{Sec.}{Secs.}
\Crefname{subsection}{Sec.}{Secs.}
\crefname{subsubsection}{Sec.}{Secs.}
\Crefname{subsubsection}{Sec.}{Secs.}
\crefname{figure}{Fig.}{Figs.}
\Crefname{figure}{Fig.}{Figs.}
\crefname{table}{Tab.}{Tabs.}
\Crefname{table}{Tab.}{Tabs.}
\crefname{equation}{Eq.}{Eqs.}
\Crefname{equation}{Eq.}{Eqs.}

\title{\bf Remembering to Reason: Structured Memory for Scene and Workflow Understanding
    in Robot-Assisted Surgery}

\author{Zhiran Zhang$^{1}$, Hu Wang$^{1}$, Yichen Wu$^{2}$, and Yutong Xie$^{1}$%
\thanks{$^{1}$Zhiran Zhang, Hu Wang, and Yutong Xie are with Mohamed bin Zayed University of Artificial Intelligence.}%
\thanks{$^{2}$Yichen Wu is with Imperial College London.}%
}

\begin{document}

\maketitle
\thispagestyle{empty}
\pagestyle{empty}

\begin{abstract}
Surgical video understanding is an important foundation for intelligent assistance in robot-assisted surgery. Similar surgical views can reflect different actions or procedural stages. Interpreting these views requires access to earlier observations, yet frame-centric prompting leaves this history unavailable. We introduce MemSurg, a framework that remembers surgical observations before reasoning over them. Its structured memory graph stores timestamped entity categories, locations, motion cues, and interaction summaries. Chronological links preserve recent context, while entity-associated links connect recurring observations across the procedure. Task-adaptive retrieval then supplies evidence-grounded prompts for scene and workflow reasoning, supporting both frozen inference and memory-conditioned supervised fine-tuning. Evaluation on robot-assisted prostatectomy and two complementary surgical benchmarks shows a 10.43-percentage-point gain over SurgRAW with GPT-4o. Memory conditioning adds 6.52 points under matched Qwen3-VL-8B fine-tuning, and an inference-time plug-in adds 7.04 points to EndoChat. These results support structured memory as a reusable source of procedural evidence for surgical video understanding. Our code will be publicly released.
\end{abstract}

\section{Introduction}
\label{sec:introduction}

Surgical video understanding is an important foundation for intelligent assistance in robot-assisted surgery. During long and demanding procedures, assistance systems could help maintain situational awareness by interpreting instruments, anatomy, and the progress of an intervention~\cite{maierhein2017surgical,hashimoto2018ai,reijmerink2024fatigue}. The surgical video provides a continuous record of these elements and their interactions. Making use of this record requires more than recognizing what appears in an image. An instrument's role depends on how it is used, and the meaning of a tissue configuration may depend on earlier manipulations. Useful understanding must therefore connect the current scene to the procedure that produced it.

\begin{figure}[t]
\centering
\includegraphics[width=\linewidth]{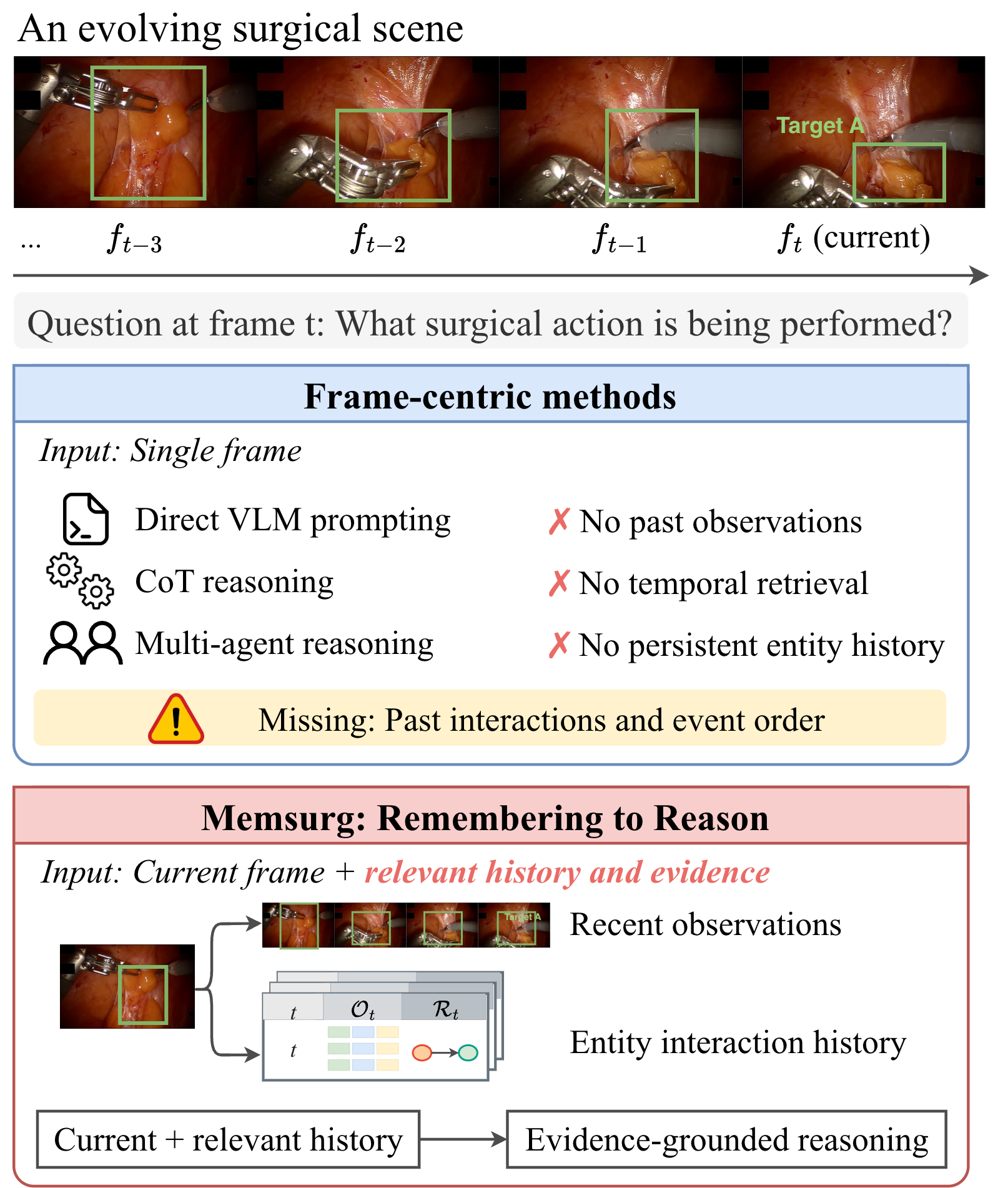}
\vspace{-10pt}
\caption{\textbf{Challenges in frame-centric surgical video understanding.} Earlier interactions and event order help explain the current scene but remain unavailable from a single frame. MemSurg retrieves this context from structured memory for evidence-grounded reasoning.}
\label{fig:motivation}
\vspace{-15pt}
\end{figure}

Language-based interfaces offer a flexible way to ask about this evolving scene. Surgical video analysis has progressed from temporal workflow recognition~\cite{jin2018svrcnet} to visual question answering and scene-graph conditioning~\cite{seenivasan2022surgicalvqavisualquestionanswering,yuan2024scenegraphvqa}. Advances in language modeling and multimodal pretraining provide broader foundations for answering questions beyond a fixed recognition task~\cite{bai2025qwen3vl}. In medical applications, however, fluent answers can still contain unsupported content~\cite{hartsock2024vlm,asgari2025clinical}. For surgical video, grounding an answer requires both interpreting visible evidence and recovering earlier observations that explain the current scene.

The availability of that earlier evidence remains a limitation of frame-centric reasoning. Chain-of-thought (CoT) and multi-agent prompting organize how a model inspects an image and compares candidate answers~\cite{Xu_2025_ICCV,low2026surgraw}. When the input contains only the current frame, these procedures cannot consult interactions observed earlier in the operation. This matters when an instrument reappears after occlusion, a target changes after manipulation, or similar views occur at different workflow stages. The required history also varies with the question: recent observations may clarify a local action, whereas earlier encounters with a recurring entity may inform procedural state. A fixed temporal window retains nearby observations even when the useful evidence is dispersed across the procedure. Memory must therefore support different temporal scopes. As illustrated in \cref{fig:motivation}, better reasoning requires access to the observations that give the current image its context.

We address this need through \textbf{Remembering to Reason}: retaining surgical observations in a form that can be retrieved when a question arises. Our framework, \textbf{MemSurg}, converts sampled video into a structured surgical memory graph. Each timestamped node records entity categories, locations, motion cues, and a semantic interaction summary. Chronological edges connect consecutive observations, while logical edges associate recurring entities across time. Together, they preserve procedural order and provide routes to entity-associated history within the same memory. The graph separates what was observed from how those observations are accessed for a later query.

MemSurg retrieves different parts of this history for scene and workflow questions. Frame-wise queries access recent chronological observations; workflow queries access earlier records associated with entities visible at the queried timestamp. Selected records are ordered by time and combined with the current image and instructions for verification, cross-referencing, and candidate elimination. This evidence interface supports frozen-VLM prompting, memory-conditioned supervised fine-tuning (SFT), and inference-time augmentation of a pretrained surgical VLM. Here, evidence grounding means conditioning answers on explicit visual observations and retrieved records. Memory construction and evidence selection can thus be evaluated separately from the model's interpretation of that evidence.

Our contributions are threefold:
\begin{itemize}
\item We introduce a structured surgical memory that combines instance-level semantic and motion cues with chronological and entity-associated connectivity, making procedural history available to subsequent questions.
\item We develop task-adaptive retrieval and evidence composition for scene and workflow reasoning. One memory interface supports frozen inference, memory-conditioned SFT, and integration with a surgical VLM.
\item We evaluate MemSurg on robotic prostatectomy and two complementary surgical benchmarks. With \mbox{GPT-4o}~\cite{openai2024gpt4o}, MemSurg-base improves accuracy by 10.43 percentage points over \mbox{SurgRAW}~\cite{low2026surgraw}; matched SFT gains 6.52 points from memory conditioning. Perception, edge, and plug-in experiments examine which components support these gains.
\end{itemize}

\section{Related Work}
\label{sec:related_work}

\subsection{Surgical Video Understanding}
Surgical video understanding has expanded from specialized recognition toward representations that support multiple tasks. Temporal models capture procedural progression~\cite{jin2018svrcnet}, while visual backbones such as \mbox{DINOv2}~\cite{oquab2024dinov2learningrobustvisual} and \mbox{LemonFM}~\cite{che2025lemon} provide reusable features for perception. Language supervision extends this direction through surgical visual question answering, scene-graph conditioning, and procedure-aware pretraining~\cite{seenivasan2022surgicalvqavisualquestionanswering,yuan2024scenegraphvqa,yuan2025procedureawaresurgicalvideolanguagepretraining,bieck2025deepvlment}. These developments enrich the interpretation of surgical images and make their content accessible through questions. They also supply the visual and semantic building blocks for a memory of the procedure. The choice of representation does not by itself determine which past observations should be retained or how their relationships should guide retrieval.

Adapting language and multimodal models to medical data further improves their ability to interpret domain-specific inputs. Cross-modal pretraining, lightweight adaptation, medical fine-tuning, and multimodal retrieval pursue complementary approaches~\cite{xlip2024,telescopic2025,yang2025medllm,mmedrag2024}. Surgical-LVLM~\cite{wang2024surgicallvlm} and \mbox{EndoChat}~\cite{wang2024endochat} adapt language-based understanding to surgical questions. Our focus is how observations from a particular procedure remain available after their original frames. This complements backbone adaptation: a model can learn surgical knowledge during training and consult procedure-specific evidence at inference. We examine this combination by adding memory to \mbox{EndoChat}~\cite{wang2024endochat} without further tuning in \cref{sec:topup}.

\subsection{Memory and Retrieval-Guided CoT}
Reasoning over extended inputs depends on how useful evidence is retained and recovered. Studies of long CoT and controlled memory examine this need across reasoning steps~\cite{chen2025reasoningerasurveylong,wang2025scmenhancinglargelanguage}. A-Mem~\cite{xu2025amemagenticmemoryllm} organizes persistent agent memories; M3-Agent~\cite{long2025seeinglisteningrememberingreasoning} and Ego-R1~\cite{tian2025egor1} explore memory or retrieval for extended video reasoning. Surgical memory must accommodate deforming anatomy, similar instruments, and intermittent visibility. MemSurg associates observations through predicted categories and visual appearance, while retaining motion and interaction descriptions as evidence. Preserving temporal order remains useful when visual similarity is ambiguous. The graph retains timestamped observations and uses estimated entity identity to provide an additional route through them. Its chronological and logical links thus provide access to recent context and possible recurrences of an entity.

Retrieved evidence must also be incorporated into the reasoning process. CoT organizes intermediate inference~\cite{wei2023chainofthoughtpromptingelicitsreasoning,Xu_2025_ICCV}, and retrieval-augmented generation supplies external evidence~\cite{liu2025ragbiomedicine}. \mbox{SurgRAW}~\cite{low2026surgraw} brings task-specific CoT and multi-agent coordination to surgical video analysis. MemSurg connects such reasoning instructions to a structured record of the procedure itself. Task-adaptive retrieval selects recent observations for frame-wise questions and entity-associated history for workflow questions. The same stored nodes therefore support both temporal scopes through a common evidence-grounded prompt interface.

\section{Methodology}
\label{sec:method}

\begin{figure*}[t]
\vspace*{6pt}
\centering
\includegraphics[width=\textwidth]{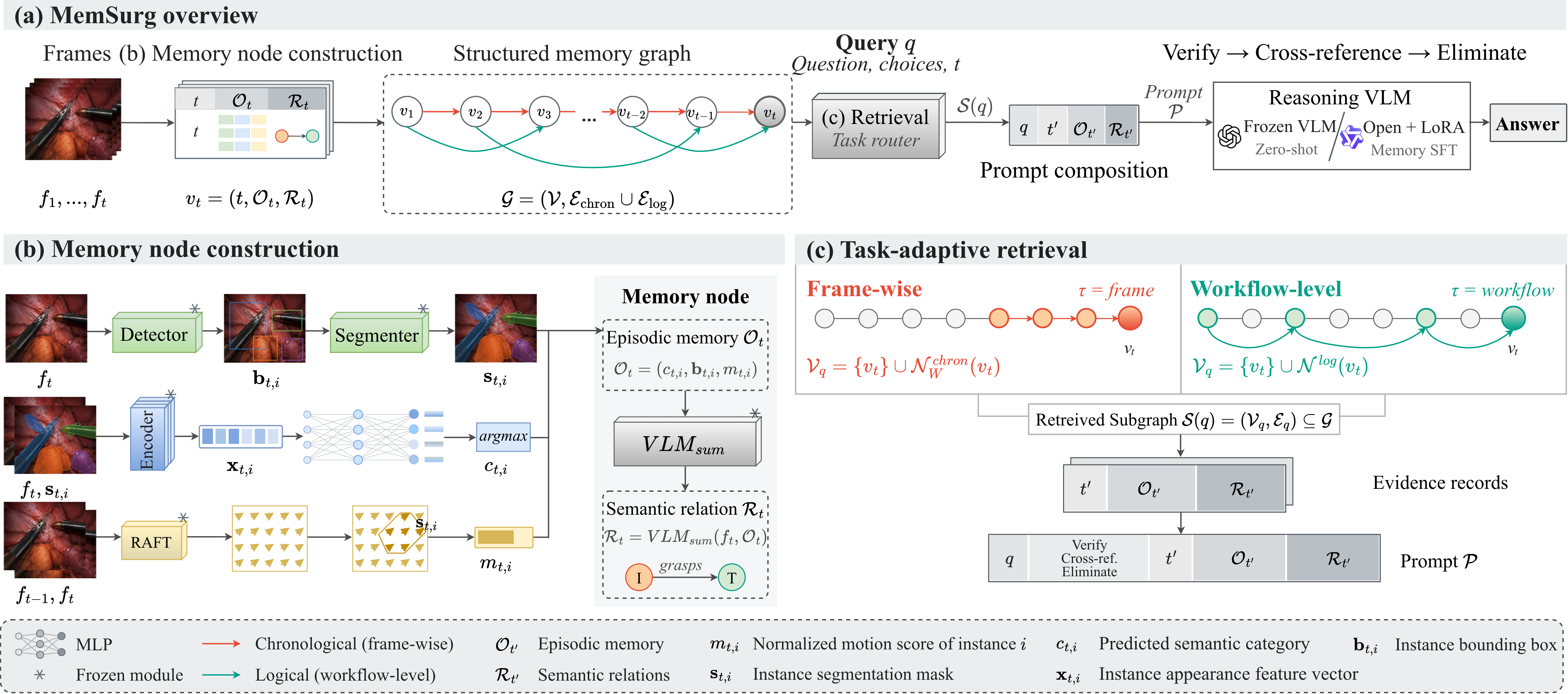}
\caption{\textbf{Overview of MemSurg.} (a) Sampled frames form timestamped memory nodes connected by chronological and entity-associated logical edges. Retrieved evidence conditions a frozen or LoRA-tuned~\cite{hu2022lora} reasoning VLM. (b) Grounding DINO~\cite{liu2024groundingdinomarryingdino} and \mbox{MedSAM}~\cite{ma2024segment} localize entities. A frozen visual encoder and trainable MLP classify instances, while mask-pooled RAFT~\cite{teed2020raft} optical flow provides motion scores. Categories, boxes, and motion form episodic memory $\mathcal O_t$; a frozen VLM summarizes interactions as $\mathcal R_t$. (c) Frame-wise queries retrieve recent chronological neighbors; workflow queries retrieve earlier entity-associated observations. Selected nodes are ordered by time and combined with the current frame, question, and instructions to verify evidence, cross-reference observations, and eliminate candidate answers.}
\label{fig2}
\vspace{-10pt}
\end{figure*}

Interpreting a surgical frame often requires observations made earlier in the same procedure. MemSurg stores this history in a structured memory graph and retrieves evidence for each question (\cref{fig2}). Given video frames $\mathcal{X}=\{f_1,\ldots,f_T\}$ sampled at 1 fps, MemSurg constructs one node per timestamp. A query $q$ specifies a timestamp $t$, a task type $\tau\in\{\mathrm{frame},\mathrm{workflow}\}$, a question, and candidate answers. The task type determines the temporal scope of retrieval.

\subsection{Datasets and Question Construction}
\label{sec:datasets}

We study robot-assisted surgical understanding through GraSP~\cite{ayobi2024pixelwise} and use two complementary settings to examine the same memory design across procedures. GraSP~\cite{ayobi2024pixelwise} contains robot-assisted radical prostatectomy with scene and workflow annotations. PitVis-2023~\cite{das2025pitvis} covers endoscopic pituitary surgery with instrument and step labels. CholecT50~\cite{nwoye2022rendezvous,nwoye2022cholect50} contains 50 laparoscopic cholecystectomy videos with instrument--verb--target triplet and phase annotations. Together, they provide a direct robotic-surgery evaluation and comparisons under different surgical and visual conditions.

To evaluate scene and workflow understanding through a common interface, we construct single-answer multiple-choice questions from the task label at each evaluated timestamp. The annotation supplies the unique correct answer, and other values in the same task-specific label space serve as incorrect options. The number of options varies across questions. PitVis-2023~\cite{das2025pitvis} and GraSP~\cite{ayobi2024pixelwise} cover available instrument/action questions and phase/step recognition or prediction questions. CholecT50~\cite{nwoye2022cholect50} covers instrument, verb, target, triplet, and phase recognition. Every method receives the same frames, questions, and options.

A reusable scene representation allows the same memory to support multiple downstream questions. The entity parser therefore predicts four generic categories: instrument, surgical target, auxiliary material, and critical structure. Spatial supervision comes exclusively from GraSP~\cite{ayobi2024pixelwise} training videos; CholecT50~\cite{nwoye2022cholect50}, PitVis-2023~\cite{das2025pitvis}, and held-out evaluation videos are excluded from parser training. These categories do not encode phase, step, verb, or triplet answers. Downstream question--answer supervision is introduced only for SFT. Accordingly, \emph{zero-shot prompting} refers to the frozen reasoning VLM, while the perception stage uses the specified spatial supervision.

\subsection{Graph and Node Construction}
\label{sec:memory_graph}

A useful memory needs observations that remain interpretable beyond their original images. Surgical scenes contain multiple instruments and anatomical structures, whose roles depend on their interactions. A global frame description can obscure which entity moved or which target was involved. We therefore represent each frame through localized entities before composing a semantic summary. This keeps category, position, and motion associated with the same instance, so they can be retained together in memory. Grounding DINO~\cite{liu2024groundingdinomarryingdino} proposes boxes, and \mbox{MedSAM}~\cite{ma2024segment} refines them into masks. For instance $i$ at time $t$, this produces a box $\mathbf b_{t,i}$ and mask $\mathbf s_{t,i}$. A frozen encoder extracts a mask-pooled appearance feature $\mathbf x_{t,i}$, which a two-layer MLP maps to the generic category $c_{t,i}$. MemSurg-base uses \mbox{DINOv2}~\cite{oquab2024dinov2learningrobustvisual}; the \mbox{LemonFM}~\cite{che2025lemon} variant substitutes a surgically pretrained encoder and retrains only the MLP. Both encoders supply the same entity fields to memory.

Each sampled scene is stored in a timestamped node that combines entity observations with a semantic description of their interaction:
\begin{equation}
\begin{split}
v_t&=(t,\mathcal O_t,\mathcal R_t),\\
\mathcal O_t&=\{(c_{t,i},\mathbf b_{t,i},m_{t,i})\}_{i=1}^{n_t},\\
\mathcal R_t&=\mathrm{VLM}_{\mathrm{sum}}(f_t,\mathcal O_t).
\end{split}
\label{eq:node}
\end{equation}
Motion complements appearance when describing an entity's activity within the scene. RAFT~\cite{teed2020raft} estimates optical flow between adjacent sampled frames. We average its magnitude within each instance mask and apply video-wise min--max normalization to obtain $m_{t,i}\in[0,1]$. This score indicates image motion; tissue contact or a surgical action still requires other visual evidence. Normalization is performed offline; the reported pipeline is therefore retrospective even though historical retrieval is restricted to earlier timestamps.

The episodic record $\mathcal O_t$ preserves what entities are present, where they appear, and how much they move. The relation summary $\mathcal R_t$ adds interactions, such as an instrument engaging a target, that help interpret these observations together. Masks are used for feature pooling and motion aggregation, while appearance features support classification and entity association. These intermediate representations are excluded from the textual episodic record. This keeps the reasoning evidence compact while preserving the information needed to index related observations.

Semantic summaries provide a higher-level interpretation of the visual evidence and depend on the model that produces them. In frozen-VLM configurations, one backbone generates both relations and answers; changing the reasoning backbone also changes the summaries. For MemSurg-SFT, the frozen base model generates summaries once before training, and they remain fixed during LoRA~\cite{hu2022lora} optimization. The configurations share the memory schema and construction procedure while retaining their model-generated content.

Isolated records preserve individual scenes, but reasoning across a procedure also requires connections between them. We organize the nodes as $\mathcal G=(\mathcal V,\mathcal E_{\mathrm{chron}}\cup\mathcal E_{\mathrm{log}})$, where $\mathcal V=\{v_1,\ldots,v_T\}$, using two complementary edge types. Chronological edges connect consecutive timestamps, $(v_{t-1},v_t)$, and preserve local procedural order. They allow a current observation to be considered alongside the events immediately preceding it.

Logical edges follow recurring entities rather than consecutive frames, allowing retrieval to reach earlier observations separated by long intervals. For a current instance $e_{t,i}$ and an earlier candidate $e_{t',j}$, association uses
\begin{equation}
s(e_{t,i},e_{t',j})=
0.3\,\mathbb I[c_{t,i}=c_{t',j}]
+0.7\cos(\mathbf x_{t,i},\mathbf x_{t',j}).
\label{eq:match}
\end{equation}
The current instance inherits the identity of the highest-scoring earlier instance when the score is at least $0.6$; otherwise it receives a new identity. These parameters are shared across datasets. Logical edges connect consecutive occurrences of each inferred identity. Such identities are estimated from generic categories and appearance, so occlusion, deformation, and visually similar instruments can introduce association errors.

Both connections access the same observations, allowing task-specific evidence selection within a fixed memory representation. A timestamp index provides direct access to the queried node, and an entity index records prior occurrence times for each inferred identity. Retrieval can therefore follow either recent procedural history or the history of a recurring entity without requiring the entire video record.

\subsection{Evidence-Grounded Reasoning}
\label{sec:reasoning}

Different questions require different parts of the remembered procedure. Frame-wise interpretation benefits from recent context, whereas workflow questions may depend on earlier interactions involving the entities now in view. For a frame-wise query, $\mathcal N_W^{\mathrm{chron}}(v_t)$ contains the preceding $W=50$ nodes, or all available preceding nodes near the start of a video. For a workflow-level query, $\mathcal N^{\mathrm{log}}(v_t)$ contains earlier nodes associated with any inferred identity visible at $t$. We select
\begin{equation}
\mathcal V_q=\{v_t\}\cup
\begin{cases}
\mathcal N_{50}^{\mathrm{chron}}(v_t), & \tau=\mathrm{frame},\\
\mathcal N^{\mathrm{log}}(v_t), & \tau=\mathrm{workflow}.
\end{cases}
\label{eq:retrieval}
\end{equation}
The corresponding edges of the selected type define the retrieved subgraph $\mathcal S(q)\subseteq\mathcal G$, with every historical node satisfying $t'<t$. Including the current node, frame-wise retrieval contains at most 51 nodes. Workflow retrieval follows entity occurrence history without a fixed node cap; an entity that persists through the video can consequently produce a long context.

This routing gives the two edge types distinct roles in reasoning. Recent observations help interpret an action through transient occlusion or appearance changes, while entity-associated records recover interactions that may clarify the current procedural state. The router uses the provided task type, with no learned query-relevance score. We examine this design by changing the retrieval edges in \cref{sec:abl_edge}.

To make the selected history available to the reasoning VLM, a deterministic template serializes $(t',\mathcal O_{t'},\mathcal R_{t'})$ in increasing timestamp order. It then appends the question, candidate answers, queried timestamp, task type, and reasoning instructions. The current frame $f_t$ remains a visual input, so the model can combine remembered observations with the image's fine-grained appearance. The final prediction is
\begin{equation}
a=\mathrm{VLM}\big(f_t,\mathrm{Compose}(\mathcal S(q),q)\big).
\label{eq:answer}
\end{equation}
The graph determines which records are retrieved; the prompt presents their ordered contents without separately verbalizing the edges. Thus, the VLM receives timestamped evidence rather than a complete textual encoding of graph topology. Retrieval is followed directly by serialization, with no learned re-ranking or additional compression.

Our prompt adapts three operations from \mbox{SurgRAW}~\cite{low2026surgraw} to use retrieved evidence. \emph{Verification} examines the current image and entity record. \emph{Cross-referencing} compares that interpretation with recent observations for frame-wise tasks or distributed entity-associated history for workflow tasks. \emph{Elimination} rejects candidate answers incompatible with the available evidence before selecting an answer. One VLM performs all three operations through the shared prompt, using the retrieved records as context for its decision.

Memory and the current image play complementary roles in this process. A generic instrument label and motion score can identify a moving tool, while earlier relation summaries provide context for its interaction with a target. The VLM combines this context with the image to resolve fine-grained distinctions between named instruments or actions. Because stored observations can be imperfect, the prompt prioritizes clear current visual evidence when records conflict. We use \emph{evidence-grounded} to describe this explicit conditioning on retrieved observations.

The same evidence interface can be used directly or incorporated into supervised adaptation. In zero-shot prompting, the reasoning VLM remains frozen and receives no downstream answer supervision. MemSurg-SFT conditions \mbox{Qwen3-VL-8B}~\cite{bai2025qwen3vl} on the current frame, question, and retrieved records from training videos. Cross-entropy is applied only to tokens encoding the final correct option or answer; prompt tokens and intermediate reasoning text are excluded from the loss. LoRA~\cite{hu2022lora} updates the reasoning model while the base weights, vision encoder, and memory-construction modules remain fixed.

To assess the contribution of memory under supervision, the memory-free SFT control uses the same backbone, questions, answers, LoRA~\cite{hu2022lora} configuration, and optimization schedule. It omits retrieved records during both training and inference. This matched comparison tests whether answer supervision and remembered evidence remain complementary. Adding the same evidence interface to a pretrained surgical VLM provides a separate test of its usefulness without further model adaptation.

\begin{figure}[t]
\vspace*{6pt}
\centering
\includegraphics[width=\columnwidth]{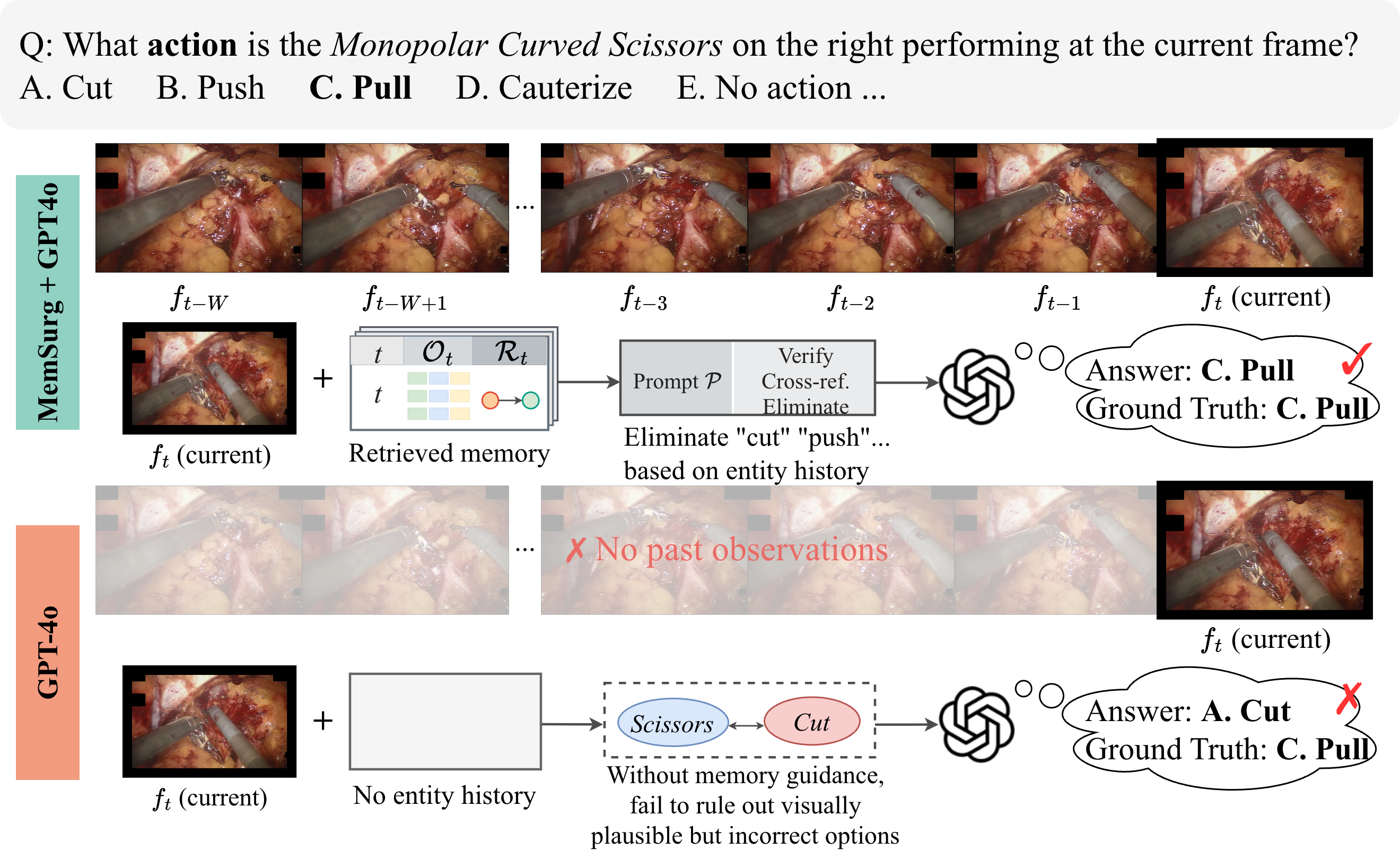}
\caption{\textbf{Memory-guided zero-shot reasoning.} The same frozen \mbox{GPT-4o}~\cite{openai2024gpt4o} answers an identical current-frame MCQ with and without MemSurg. Retrieved chronological records support pulling over visually plausible cutting in this illustration. Historical frames show the memory source; only the current image and structured text enter the MemSurg reasoning prompt.}
\label{fig3}
\vspace{-10pt}
\end{figure}

\section{Experiments}
\label{sec:experiments}

We evaluate frame-wise and workflow-level questions to assess memory under different temporal demands. Frozen inference, matched supervised adaptation, and an inference-time plug-in test its utility across reasoning models, while component ablations examine memory content and retrieval connections. We report accuracy on the common multiple-choice question set. Overall accuracy is the fraction of correctly answered questions across all datasets, with equal weight per question and no dataset- or task-level reweighting. Task averages summarize the corresponding frame-wise or workflow-level scores across datasets. All improvements are reported as absolute percentage-point differences.

\subsection{Implementation}
\label{sec:implementation}

\minisection{Perception and entity training.}
The perception setup keeps the memory interface consistent while allowing the entity representation to be compared. Grounding DINO~\cite{liu2024groundingdinomarryingdino}, \mbox{MedSAM}~\cite{ma2024segment}, RAFT~\cite{teed2020raft}, and the visual encoder remain frozen. Only the MLP is trained, using spatially annotated GraSP~\cite{ayobi2024pixelwise} training instances mapped to the four generic categories. CholecT50~\cite{nwoye2022cholect50}, PitVis-2023~\cite{das2025pitvis}, and held-out GraSP~\cite{ayobi2024pixelwise} videos are excluded. The default visual encoder is \mbox{DINOv2}~\cite{oquab2024dinov2learningrobustvisual} ViT-L/14.

The two-layer MLP uses 512 hidden units, ReLU, and dropout of 0.1. We train it with AdamW for 20 epochs, using batch size 256, a learning rate of $10^{-3}$ with cosine decay, and weight decay of 0.01.

\minisection{Reasoning configurations.}
For frozen inference, we use \mbox{GPT-4o}~\cite{openai2024gpt4o} or \mbox{Qwen3-VL-30B-A3B}~\cite{bai2025qwen3vl} with the same retrieval interface. MemSurg-base uses \mbox{DINOv2}~\cite{oquab2024dinov2learningrobustvisual} for memory construction. MemSurg-LemonFM~\cite{che2025lemon} changes the visual encoder while retaining \mbox{GPT-4o}~\cite{openai2024gpt4o}, the retrieval rule, and the reasoning template. The generic CoT comparison applies a staged prompt based on LLaVA-CoT~\cite{Xu_2025_ICCV} to \mbox{GPT-4o}~\cite{openai2024gpt4o}; it evaluates this prompting strategy rather than the original checkpoint. \mbox{SurgRAW}~\cite{low2026surgraw} provides the surgical multi-agent baseline. We report frozen and fine-tuned configurations separately because they use different backbones and supervision.

MemSurg-SFT adapts \mbox{Qwen3-VL-8B}~\cite{bai2025qwen3vl} using questions from the training videos of all three datasets. The base model generates semantic relations once before optimization. LoRA~\cite{hu2022lora} uses rank 16, scaling 32, and dropout of 0.05, with the vision encoder and base weights frozen. AdamW training runs for three epochs with a learning rate of $2\times10^{-4}$, weight decay of 0.01, cosine decay, 3\% warmup, and effective batch size 64. The memory-free control shares these data and settings. 

\minisection{Native perception quality.}
We evaluate instrument localization and segmentation in the native pipeline on held-out GraSP~\cite{ayobi2024pixelwise} frames. Without detector fine-tuning, Grounding DINO~\cite{liu2024groundingdinomarryingdino} achieves 61\% box AP$_{50}$ and 80\% recall. Frozen \mbox{MedSAM}~\cite{ma2024segment}, prompted with these predicted detector boxes rather than ground-truth boxes, achieves 77\% mean Dice and 63\% mean IoU. Bounding boxes provide the spatial fields stored in episodic records, while masks define the regions used for appearance-feature pooling and optical-flow aggregation.

\subsection{Results}
\label{sec:main_results}

\begin{table*}[t]
\vspace*{6pt}
\caption{\textbf{Surgical video understanding across three benchmarks.} Accuracy (\%) for frame-wise and workflow-level questions. Overall pools all evaluated questions; Task Avg. averages dataset scores within each task group. Bold marks the best result within each evaluation setting.}
\label{tab1}
\centering
\footnotesize
\setlength{\tabcolsep}{2pt}
\begin{tabularx}{\textwidth}{>{\raggedright\arraybackslash}p{0.34\textwidth}*{9}{>{\centering\arraybackslash}X}}
\toprule
\multirow{2}{*}{Method} & \multirow{2}{*}{Overall} &
\multicolumn{2}{c}{Task Avg.} &
\multicolumn{2}{c}{PitVis-2023~\cite{das2025pitvis}} &
\multicolumn{2}{c}{GraSP~\cite{ayobi2024pixelwise}} &
\multicolumn{2}{c}{CholecT50~\cite{nwoye2022cholect50}} \\
\cmidrule(lr){3-4}\cmidrule(lr){5-6}\cmidrule(lr){7-8}\cmidrule(lr){9-10}
& & Frame & Workflow & Frame & Workflow & Frame & Workflow & Frame & Workflow \\
\midrule
\multicolumn{10}{l}{\textit{Prompting-based methods}} \\
\rowcolor{gray!15}\mbox{GPT-4o} (MCQ)~\cite{openai2024gpt4o}
& 41.15 & 39.97 & 42.34 & 40.93 & 43.46 & 41.44 & 45.45 & 37.54 & 38.11 \\
\mbox{Qwen3-VL-30B-A3B} (MCQ)~\cite{bai2025qwen3vl}
& 28.50 & 28.38 & 28.53 & 28.85 & 28.30 & 30.06 & 30.20 & 26.23 & 27.09 \\
\rowcolor{gray!15}CoT prompt (\mbox{GPT-4o}~\cite{openai2024gpt4o})~\cite{Xu_2025_ICCV}
& 38.75 & 38.21 & 39.29 & 39.00 & 40.21 & 40.66 & 42.08 & 34.97 & 35.58 \\
\mbox{SurgRAW} (\mbox{GPT-4o}~\cite{openai2024gpt4o})~\cite{low2026surgraw}
& 53.14 & 50.49 & 55.79 & 50.33 & 56.93 & 52.34 & 57.44 & 48.80 & 53.00 \\
\rowcolor{gray!15}MemSurg-base (\mbox{Qwen3-VL-30B-A3B}~\cite{bai2025qwen3vl})
& 38.01 & 34.70 & 41.32 & 34.90 & 41.61 & 36.83 & 43.66 & 32.37 & 38.69 \\
MemSurg-base (\mbox{GPT-4o}~\cite{openai2024gpt4o})
& 63.57 & 61.89 & 65.25 & 62.39 & 65.90 & 63.99 & 67.40 & 59.29 & 62.45 \\
\rowcolor{c1!30}\textbf{MemSurg-LemonFM~\cite{che2025lemon} (\mbox{GPT-4o}~\cite{openai2024gpt4o})}
& \textbf{73.73} & \textbf{74.96} & \textbf{72.50}
& \textbf{75.66} & \textbf{73.22} & \textbf{76.68} & \textbf{73.91} & \textbf{72.54} & \textbf{70.37} \\
\midrule
\multicolumn{10}{l}{\textit{Fine-tuned \mbox{Qwen3-VL-8B}~\cite{bai2025qwen3vl} methods}} \\
\rowcolor{gray!15}\mbox{Qwen3-VL-8B}~\cite{bai2025qwen3vl} SFT (no memory)
& 47.20 & 45.62 & 48.77 & 45.24 & 49.14 & 48.75 & 50.69 & 42.87 & 46.48 \\
\rowcolor{c1!30}\textbf{MemSurg-SFT (\mbox{Qwen3-VL-8B}~\cite{bai2025qwen3vl})}
& \textbf{53.72} & \textbf{52.63} & \textbf{54.81}
& \textbf{53.52} & \textbf{55.47} & \textbf{54.87} & \textbf{56.37} & \textbf{49.50} & \textbf{52.59} \\
\bottomrule
\end{tabularx}
\vspace{-6pt}
\end{table*}

\minisection{Frozen inference.}
Stored observations benefit both scene and workflow reasoning. MemSurg-base with \mbox{GPT-4o}~\cite{openai2024gpt4o} reaches 63.57\% overall accuracy, compared with 53.14\% for \mbox{SurgRAW}~\cite{low2026surgraw} and 41.15\% for direct multiple-choice prompting (\cref{tab1}). The gains over the surgical reasoning baseline are 11.40 points on frame-wise questions and 9.46 points on workflow-level questions. Remembered evidence thus complements structured surgical reasoning prompts for both scene interpretation and procedural understanding.

\Cref{fig3} illustrates why remembered context matters for action recognition. Instrument appearance alone can make an incorrect option visually plausible. Retrieved observations supply temporal evidence unavailable in the current frame, helping the same frozen model reject candidate answers inconsistent with the preceding interaction.

The benefit carries through to the robot-assisted benchmark at the center of this study. On GraSP~\cite{ayobi2024pixelwise}, MemSurg-base reaches 63.99\% frame-wise and 67.40\% workflow-level accuracy, exceeding \mbox{SurgRAW}~\cite{low2026surgraw} by 11.65 and 9.96 points. LemonFM-based memory~\cite{che2025lemon} raises these scores to 76.68\% and 73.91\%. Gains on PitVis-2023~\cite{das2025pitvis} and CholecT50~\cite{nwoye2022cholect50} further support using the same memory interface across the three evaluated procedures.

Changing the frozen reasoning backbone preserves the value of the remembered context. \mbox{Qwen3-VL-30B-A3B}~\cite{bai2025qwen3vl} improves from 28.50\% with direct prompting to 38.01\% with MemSurg-base, a gain of 9.51 points. Frame-wise tasks improve by 6.32 points and workflow tasks by 12.79 points. Both tested backbones therefore benefit from memory, with a larger workflow gain in this configuration.

\minisection{Fine-tuning and plug-in use.}
Memory also remains useful when the reasoning model has access to downstream answer supervision. Under matched \mbox{Qwen3-VL-8B}~\cite{bai2025qwen3vl} fine-tuning, MemSurg-SFT increases overall accuracy from 47.20\% to 53.72\%. Frame-wise and workflow-level accuracy improve by 7.01 and 6.04 points, respectively, and every dataset--task cell improves. These matched comparisons support retrieved evidence as a complement to supervised learning.

\phantomsection\label{sec:topup}
A surgical VLM that has already been trained offers a further test of memory's role. We prepend retrieved evidence to off-the-shelf \mbox{EndoChat}~\cite{wang2024endochat}, without additional fine-tuning or access to its training data. Overall accuracy increases from 49.55\% to 56.59\% (\cref{tab:tab_topup}). The gain is larger for workflow questions (9.47 points) than for frame-wise questions (4.61 points). An external record of procedural observations thus complements the pretrained model at inference.

\begin{table}[!tbp]
\centering
\caption{\textbf{Memory as an inference-time plug-in.} Accuracy (\%) of the same pretrained \mbox{EndoChat}~\cite{wang2024endochat} with and without retrieved memory. No additional fine-tuning is performed.}
\label{tab:tab_topup}
\footnotesize
\setlength{\tabcolsep}{4pt}
\begin{tabularx}{\columnwidth}{>{\raggedright\arraybackslash}Xccc}
\toprule
Method & Overall & Frame & Workflow \\
\midrule
\rowcolor{gray!15}\mbox{EndoChat}~\cite{wang2024endochat} (off-the-shelf) & 49.55& 51.67& 47.43\\
\rowcolor{c1!30}\textbf{EndoChat~\cite{wang2024endochat} (memory)} & 56.59& 56.28& 56.90\\
\bottomrule
\end{tabularx}
\vspace{-10pt}
\end{table}

\subsection{Ablation}
\label{sec:ablation}

\minisection{Visual backbone.}
\phantomsection\label{sec:abl_backbone}
A useful memory design should exploit improved perception while retaining the same interface for retrieval and reasoning. Building on the spatial grounding assessed in the ``Native perception quality'' paragraph of \cref{sec:implementation}, we compare the default \mbox{DINOv2}~\cite{oquab2024dinov2learningrobustvisual} encoder with the surgically pretrained \mbox{LemonFM}~\cite{che2025lemon}. Each encoder remains frozen, and its MLP head is trained with the settings in \cref{sec:implementation}. We keep \mbox{GPT-4o}~\cite{openai2024gpt4o}, the memory fields, retrieval rules, and reasoning prompt fixed, with instance recognition (IR) category fields and optical flow (OP) motion scores in the prompt.

\begin{figure}[!t]
\vspace*{6pt}
\centering
\includegraphics[width=\columnwidth]{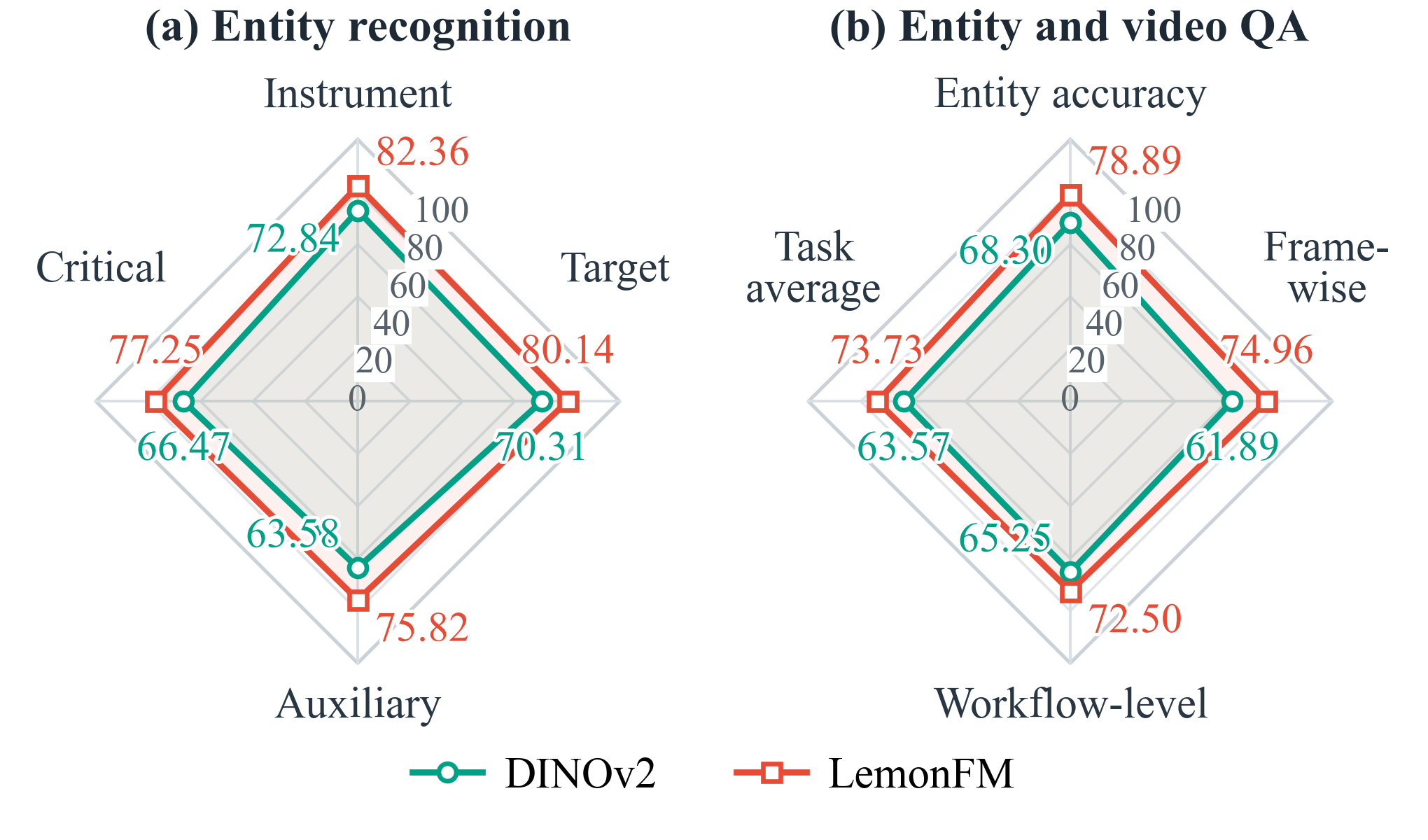}
\vspace{-18pt}
\caption{\textbf{Visual backbone ablation.} Frozen \mbox{DINOv2}~\cite{oquab2024dinov2learningrobustvisual} and \mbox{LemonFM}~\cite{che2025lemon} encoders are compared under matched MemSurg-base settings, with category and motion evidence enabled. (a) Recognition accuracy for the four entity categories. (b) Aggregate entity accuracy and downstream QA accuracy. All scores are percentages.}
\label{fig4}
\end{figure}

The native configuration already produces useful entity representations: \mbox{DINOv2}~\cite{oquab2024dinov2learningrobustvisual} achieves 68.30\% entity accuracy and 63.57\% downstream task-average accuracy (\cref{fig4}). Replacing it with \mbox{LemonFM}~\cite{che2025lemon} improves recognition in all four entity categories and raises aggregate entity accuracy to 78.89\%. Auxiliary materials show the largest category gain ($+12.24$ points). The stronger representation also raises frame-wise accuracy from 61.89\% to 74.96\% and workflow-level accuracy from 65.25\% to 72.50\%, increasing the task average to 73.73\%.

These gains show that the same structured memory can translate a stronger visual representation into better surgical video understanding. The improvement extends a working native pipeline while preserving its evidence interface, supporting the design and its potential to benefit from advances in surgical perception. Since the encoder supplies features for both classification and association, this comparison evaluates their combined contribution to memory construction.

\minisection{Feature extraction.}
\phantomsection\label{sec:abl_feature}
To understand which stored observations help reasoning, we vary the evidence presented in the prompt while keeping entity association fixed. The IR ablation removes only the category fields from the serialized prompt. Categories are still computed for association, so the graph and retrieval rule remain unchanged. This configuration tests the value of explicitly exposing entity categories to the reasoning VLM, rather than changing which observations are linked or retrieved.

Explicit category fields provide the larger individual contribution, while motion fields become more useful in their presence. With \mbox{GPT-4o}~\cite{openai2024gpt4o} and \mbox{DINOv2}~\cite{oquab2024dinov2learningrobustvisual}, the configuration without explicit IR or OP fields reaches 46.38\% average accuracy (\cref{tab:tab_feat}). IR alone reaches 56.92\%, compared with 48.50\% for OP alone. Adding OP to IR yields a further 6.65-point improvement, reaching 63.57\%. Motion contributes more when the prompt identifies the entities to which it belongs.

The same complementarity appears in both temporal scopes. Adding OP to IR raises frame-wise accuracy from 55.63\% to 61.89\% and workflow accuracy from 58.20\% to 65.25\%. Motion can therefore provide useful context within recent observations as well as longer entity histories. These single-run comparisons support retaining both cues; repeated runs would characterize the variability of the gains.

\begin{table}[!tbp]
\centering
\caption{\textbf{Contributions of entity and motion evidence.} MemSurg-base accuracy (\%). IR: category fields in the prompt; OP: optical-flow motion scores. Entity association and retrieval remain fixed. Bold marks the best scores.}
\label{tab:tab_feat}
\footnotesize
\setlength{\tabcolsep}{5pt}
\begin{tabularx}{\columnwidth}{YYYYY}
\toprule
IR & OP & Frame & Workflow & Avg. \\
\midrule
\rowcolor{gray!15}-- & -- & 48.95 & 43.80 & 46.38 \\
$\checkmark$ & -- & 55.63 & 58.20 & 56.92 \\
\rowcolor{gray!15}-- & $\checkmark$ & 51.62 & 45.37 & 48.50 \\
\rowcolor{c1!30}$\checkmark$ & $\checkmark$ & \textbf{61.89} & \textbf{65.25} & \textbf{63.57} \\
\bottomrule
\end{tabularx}
\vspace{-10pt}
\end{table}

\minisection{Edge use.}
\phantomsection\label{sec:abl_edge}
Useful memory also depends on selecting the history appropriate to the question. We retain \mbox{GPT-4o}~\cite{openai2024gpt4o}, \mbox{DINOv2}~\cite{oquab2024dinov2learningrobustvisual}, IR, OP, and the prompt template while changing only the retrieval edge type (\cref{tab:tab_edge}). Replacing chronological retrieval with logical retrieval for frame-wise questions reduces accuracy from 61.89\% to 52.28\%. Replacing logical retrieval with chronological retrieval for workflow questions reduces accuracy from 65.25\% to 61.79\%. The two task groups thus favor different histories: recent context for scene interpretation and recurring-entity observations for workflow understanding.

Simply supplying both histories does not recover the task-adaptive result. The Chronological + Logical setting retrieves the union of the queried node, its chronological neighbors, and its logical neighbors, rather than the complete graph. It reaches 55.67\% frame-wise and 62.04\% workflow accuracy, trailing the proposed setting by 6.22 and 3.21 points. More context is therefore not automatically more useful. Although this comparison does not separately control context length, it supports matching evidence selection to the question instead of combining the two neighborhoods indiscriminately.

\begin{table}[!tbp]
\vspace*{6pt}
\centering
\caption{\textbf{Task-adaptive memory retrieval.} MemSurg-base accuracy (\%). Task-adaptive retrieval uses chronological edges for frame-wise queries and logical edges for workflow queries. Chronological + Logical combines both neighborhoods; -- indicates not evaluated.}
\label{tab:tab_edge}
\footnotesize
\setlength{\tabcolsep}{3.5pt}
\begin{tabularx}{\columnwidth}{l>{\raggedright\arraybackslash}Xcc}
\toprule
Task & Edge type & Frame & Workflow \\
\midrule
\rowcolor{gray!15}Frame & Logical (mismatched) & 52.28 & -- \\
Workflow & Chronological (mismatched) & -- & 61.79 \\
\rowcolor{gray!15}Both tasks & Chronological + Logical & 55.67& 62.04\\
\rowcolor{c1!30}\multicolumn{2}{l}{\textbf{MemSurg (task-adaptive, ours)}} & \textbf{61.89} & \textbf{65.25} \\
\bottomrule
\end{tabularx}
\vspace{-10pt}
\end{table}

These experiments establish improved multiple-choice performance on retrospective surgical videos. Generative medical-video reasoning, including reinforcement-learning approaches such as MedGRPO~\cite{su2026medgrpo}, follows a different answer protocol and falls outside this comparison. Moving toward robot-assisted deployment also requires evaluating rationale faithfulness and latency. In the current pipeline, workflow histories can grow with video duration and motion normalization remains offline, making bounded retrieval and causal calibration priorities for subsequent development.
\section{Conclusion}
MemSurg puts \emph{Remembering to Reason} into practice by preserving surgical observations as structured, retrievable evidence. A common record of entities and interactions supports recent chronological retrieval for scene understanding and entity-associated retrieval for workflow understanding. Experiments on robot-assisted prostatectomy and two complementary surgical benchmarks show benefits with frozen VLMs, matched supervised fine-tuning, and an \mbox{EndoChat}~\cite{wang2024endochat} plug-in. Feature and edge ablations support the joint value of informative memory and task-appropriate evidence selection. Future work will examine causal motion calibration, bounded workflow retrieval, and evidence faithfulness beyond multiple-choice answers.

\bibliographystyle{bst_file/IEEEtran_memsurg}
\bibliography{egbib}
\end{document}